\documentclass[letterpaper, 10 pt, conference]{ieeeconf}  

\IEEEoverridecommandlockouts                              

\usepackage{graphics} 
\usepackage{amsmath} 
\usepackage{tikz}
\usepackage{import}
\usepackage{pgfplots}
\usepackage{graphicx}
\graphicspath{{./Imgs/}}
\usepackage{todonotes}
\usepackage{array}
\usepackage[utf8]{inputenc}

\title{\LARGE \bf
Learning the Subsystem of Local Planning for Autonomous Racing
}

\author{Benjamin Evans$^{1}$, Herman A. Engelbrecht$^{1}$ and Hendrik W. Jordaan$^{1}$

\thanks{$^{1}$Electrical and Electronic Engineering Department,
        Stellenbosch University, Stellenbosch, 7600, South Africa. 
        {\tt\small \{19811799, hebrecht, wjordaan\}@sun.ac.za};%
        }
}

\begin{document}

\maketitle
\thispagestyle{empty}
\pagestyle{empty}

\begin{abstract}

The problem of autonomous racing is to navigate through a race course as quickly as possible while not colliding with any obstacles.
We approach the autonomous racing problem with the added constraint of not maintaining an updated obstacle map of the environment.
Several current approaches to this problem use end-to-end learning systems where an agent replaces the entire navigation pipeline.
This paper presents a hierarchical planning architecture that combines a high level planner and path following system with a reinforcement learning agent that learns that subsystem of obstacle avoidance.

The novel ``modification planner'' uses the path follower to track the global plan and the deep reinforcement learning agent to modify the references generated by the path follower to avoid obstacles.
Importantly, our architecture does not require an updated obstacle map and only 10 laser range finders to avoid obstacles.
The modification planner is evaluated in the context of F1/$10^{th}$ autonomous racing and compared to a end-to-end learning baseline, the Follow the Gap Method and an optimisation based planner.
The results show that the modification planner can achieve faster average times compared to the baseline end-to-end planner and a 94\% success rate which is similar to the baseline.

\end{abstract}

\section{Introduction}

Autonomous racing is the problem of generating speed and steering references for a non-holonomic vehicle, that move the vehicle through a race course as quickly as possible. 
Successful approaches to autonomous racing have implemented a hierarchical approach by splitting the problem into global planning, generating a high-level plan before the mission begins, and local planning, following the global plan while avoiding obstacles \cite{betz2019software}.
Current approaches to local planning for autonomous racing, generally assume that a map of the obstacles is available or can be built \cite{betz2019software, vazquez2020optimization}.
It is difficult to simultaneously achieve the dual objective on obstacle avoidance without a map of the obstacles and following a global plan because information about the environment is high dimensional and difficult to plan with.

In this paper, we address the problem of local planning in autonomous racing without maintaining an explicit obstacle map. 
We present our novel approach which is a hybrid local planner that integrates a path follower and a deep reinforcement learning (RL) agent that modifies the steering reference generated by the path follower to accomplish the dual objective of tracking a reference path while also smoothly avoiding obstacles. 
We call our approach a reference modification planner since we learn the subsystem of obstacle avoidance by modifying the steering angle.
We evaluate our novel approach in the context of an F1/$10^{th}$ autonomous racing simulator and compare it to current approaches which include a end-to-end RL planner, the Follow the Gap Method, and an oracle optimisation based planner.
The results show the advantages of our approach, namely, it does not require a map to avoid obstacles, and it can find smoother trajectories than end-to-end approaches.

\begin{figure}
    \centering
    \def\svgwidth{0.46\textwidth}
\begingroup%
  \makeatletter%
  \providecommand\color[2][]{%
    \errmessage{(Inkscape) Color is used for the text in Inkscape, but the package 'color.sty' is not loaded}%
    \renewcommand\color[2][]{}%
  }%
  \providecommand\transparent[1]{%
    \errmessage{(Inkscape) Transparency is used (non-zero) for the text in Inkscape, but the package 'transparent.sty' is not loaded}%
    \renewcommand\transparent[1]{}%
  }%
  \providecommand\rotatebox[2]{#2}%
  \newcommand*\fsize{\dimexpr\f@size pt\relax}%
  \newcommand*\lineheight[1]{\fontsize{\fsize}{#1\fsize}\selectfont}%
  \ifx\svgwidth\undefined%
    \setlength{\unitlength}{219.08864179bp}%
    \ifx\svgscale\undefined%
      \relax%
    \else%
      \setlength{\unitlength}{\unitlength * \real{\svgscale}}%
    \fi%
  \else%
    \setlength{\unitlength}{\svgwidth}%
  \fi%
  \global\let\svgwidth\undefined%
  \global\let\svgscale\undefined%
  \makeatother%
  \begin{picture}(1,0.82864548)%
    \lineheight{1}%
    \setlength\tabcolsep{0pt}%
    \put(0,0){\includegraphics[width=\unitlength,page=1]{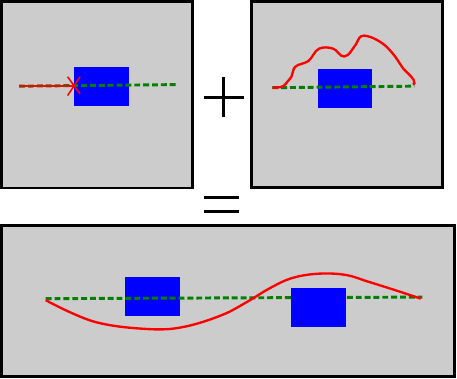}}%
    \put(0.22731354,0.52421879){\color[rgb]{0,0,0}\makebox(0,0)[t]{\lineheight{1.25}\smash{\begin{tabular}[t]{c}Path \\Follower\end{tabular}}}}%
    \put(0.74985156,0.46895727){\color[rgb]{0,0,0}\makebox(0,0)[t]{\lineheight{1.25}\smash{\begin{tabular}[t]{c}RL Agent\end{tabular}}}}%
    \put(0.50288692,0.26625075){\color[rgb]{0,0,0}\makebox(0,0)[t]{\lineheight{1.25}\smash{\begin{tabular}[t]{c}Hybrid Modification Planner\end{tabular}}}}%
  \end{picture}%
\endgroup%

    \caption{\textbf{Illustration of Hybrid Planning Approach:} The green dotted line is the global plan, the blue blocks are the obstacles and the red line is the path taken by the vehicle.
    A path following algorithm that can follow a path is combined with an RL agent that learns the subsystem of obstacle avoidance to create the modification planner.}
    \label{fig:hybrid_avoid}
\end{figure}
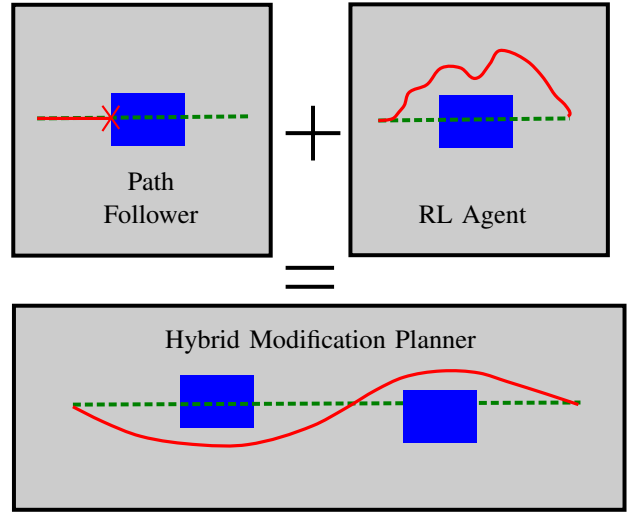

The rest of the paper is summarised as follows. 
We first discuss previous attempts to solve the local planning problem for autonomous racing in Section \ref{sec:related_work}. 
This is followed by a theoretical overview of our proposed reference modification planner in Section \ref{sec:theory}.
In Section \ref{sec:components} we explain the components that we use in building our solution and in Section \ref{sec:learning} the learning formulation for the Reinforcement Learning (RL) agent is explained.
Lastly in Sections \ref{sec:evaluation_method} and \ref{sec:results} we detail the evaluation methodology and present our results.

\section{Related Work}\label{sec:related_work}

Autonomous racing is built on the navigation problem of moving from a start to a goal location.
We study different methods of local planning for navigation and racing to develop our solution.  
We study reactive methods \cite{lumelsky1987path, sezer2012novel}, classical planning techniques \cite{autoObsAvoid_MAV, coulter1992implementation}, learning based approaches \cite{virtuaToReal, plato} and hybrid systems \cite{gao2017intention, rl_path_tracking_ugv}.

The earliest obstacle avoidance algorithms were reactive algorithms (such as ``The Bug'' algorithm) which generate steering references based on the surrounding environment at each step \cite{lumelsky1987path}.
One of the most recently proposed reactive obstacle avoidance algorithms is the ``Follow the Gap Method'' (FGM) \cite{sezer2012novel}. 
The FGM method uses a dense LiDAR scan to navigate away from the nearest obstacle into free space.
Reactive methods are not inherently able to follow a global plan and since they do not have a model of the vehicle, they are not able to plan for ahead.

Classical local planning methods aimed to improve on reactive methods by storing a map of the environment which can be used to plan a trajectory, multiple time-steps in advance.
Classical methods typical build a map using a Simultaneous Localisation and Mapping (SLAM) algorithm \cite{SLAM_paper}, and then use the map to plan and optimise a collision-free trajectory \cite{autoObsAvoid_MAV}.
The trajectory is then followed using a path tracking algorithm such as Pure Pursuit \cite{coulter1992implementation} or Model Predictive Control \cite{vazquez2020optimization}.
Classical planning and path tracking methods can generate and execute near-optimal trajectories, however, because map building is computationally expensive and slow, these solutions are not feasible for fast-moving, dynamic environments, such as autonomous racing, where all the processing must run on-board \cite{ppf_SLAM}.

Due to the hardware requirements/constraints of classical planning solutions, machine learning (ML) methods have been applied to the problem of local planning for autonomous navigation \cite{xiao2020motion}.
Neural networks, trained with reinforcement learning (RL) \cite{virtuaToReal} and imitation learning (IL) \cite{plato} have been used to associate sensor readings (camera images and LiDAR scans) directly with navigation commands.
It has been shown that agents trained end-to-end are able to learn to race autonomously and perform competitively \cite{perot2017end, vitelli2016carma}.
While using an RL agent as a local planner does not require the expensive maintenance of a map, RL agents trained end-to-end can exhibit poor long term performance and have shown to require a high-level planner in certain circumstances \cite{fred_mapless_navigation}.

Hybrid architectures that use RL agents to learn a subsystem of the planning pipeline have shown improved results over end-to-end methods \cite{xiao2020motion}.
Paths generated with A* search have been used in conjunction with a neural network as a local planner and shown to be able to execute long missions \cite{rl_path_tracking_ugv, gao2017intention}.
Learning the subsystem of trajectory planning has shown to outperform end-to-end methods \cite{weiss2020deepracing}.
We further research into learning subsystems for autonomous racing by using an RL agent to learn the local planning subsystem of obstacle avoidance for autonomous racing.
We aim to improve the performance of the vehicle by giving the agent less influence over the steering reference.
Our solution doesn't require an updated obstacle map, as with classical model-based planning solutions, and can harness the adaptability of learning techniques to form a planner which is able to smoothly avoid obstacles while maintaining a global trajectory.

\section{Theoretical Design}\label{sec:theory}

We address the problem of generating navigation references for autonomous racing that results in a reference trajectory being followed and obstacles being smoothly avoided as shown in Figure \ref{fig:hybrid_avoid}.
The solution should track a reference trajectory and deviate as necessary to safely and efficiently avoid the obstacle.
The desired behaviour is analogous to driving a car, seeing an obstacle in the road, swerving around it and then returning to the original lane.

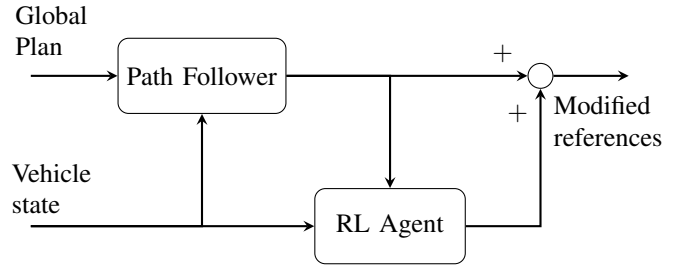
\begin{figure}[h]
    \centering
    \usetikzlibrary{shapes.geometric, arrows}

\usetikzlibrary{shapes, shadows, arrows, arrows.meta}

\tikzstyle{empty} = [rectangle, rounded corners, draw=white, minimum height = 0.2cm, minimum width=0.2cm]
\tikzstyle{block} = [rectangle, rounded corners, draw=black, minimum height = 1cm, minimum width=2cm]
\tikzstyle{mysum} = [circle, draw=black]

\tikzstyle{arrow} = [thick, ->, >=stealth]


\begin{tikzpicture}[node distance=2cm]

    \node (base) [empty] {};
    \node (b2) [empty, below of=base] {};
    \node (pf) [block, right of=base, xshift=0.4cm] {Path Follower};
    \node (nn) [block, below of=pf, xshift=2.5cm] {RL Agent};
    \node (sum) [mysum, right of=pf, xshift=2.5cm] {};
    \node (b3) [empty, right of=sum, xshift=-20] {};

    \draw [arrow] (pf) -| (nn);
    \draw [arrow] (nn) -| (sum);
    \draw [arrow] (pf) -- (sum);
    \draw [arrow] (base) -- (pf);
    \draw [arrow] (b2) -- (nn);
    \draw [arrow] (b2) -| (pf);
    \draw [arrow] (sum) -- (b3);
    
    \node[align=left] at (0.4, 0.6) {Global \\ Plan};
    \node[align=left] at (0.4, -1.5) {Vehicle \\ state};
    \node at (6.4, 0.3) {$+$};
    \node at (6.6, -0.5) {$+$};
    \node[align=left] at (7.8, -0.6) {Modified \\ references};

\end{tikzpicture}

    \caption{\textbf{Proposed Local Planner Architecture}: A path follower to track the reference path is combined in parallel with an RL agent that modifies the path follower references.}
    \label{fig:local_planner}
\end{figure}

We present a hybrid local planner that features a pure pursuit path follower in parallel with an RL agent.
Figure \ref{fig:local_planner} shows how we combine the path follower and an RL agent in parallel such that the agent can modify the path follower references to avoid obstacles.
This system is designed that the path follower maintains the reference path and the agent can deal with the uncertainty of obstacles.

The path follower uses the current location of the vehicle and a list of way-points to calculate the required steering and velocity commands to follow the global plan.
an RL agent is added in parallel to avoid obstacles by modifying the steering angle calculated by the path follower.
We train the agent by punishing it for crashing and rewarding it for progress along the global plan.
These two components in the reward signal represent the dual objectives of achieving efficient, collision-free paths.

Where previous attempts have used RL agents to replace the local planner and learn to navigate, we use an RL agent to learn to navigate as a subsystem within our local planner.
This is done with the aim of better utilising the global path and avoiding obstacles more smoothly.
The aim is that the agent will learn to swerve around obstacles where needed without modifying the behaviour of the vehicle when no obstacles are present.

\section{Preliminaries: Navigation Stack}\label{sec:components}
\begin{figure}[h]
    \centering
    \tikzstyle{empty} = [rectangle, rounded corners, draw=white, minimum height = 1cm, minimum width=2cm]
\tikzstyle{block} = [rectangle, rounded corners, draw=black, minimum height = 1cm, minimum width=2cm]
\tikzstyle{dash_block} = [rectangle, square corners, draw=black, ]

\tikzstyle{arrow} = [thick, ->, >=stealth]




\resizebox{0.46\textwidth}{!}{
    \begin{tikzpicture}[node distance=2cm]
    
        \node (Gplan) [block] {Global Planner};
        \node (map) [empty, left of=Gplan, xshift=-1cm] {Environment Map};
        \node (Lplan) [block, below of=Gplan, yshift=0.5cm] {Local Planner};
        \node (c_sys) [block, right of=Lplan, xshift=1.5cm] {Control System};
        \node (sensors) [block, left of=Lplan, xshift=-1cm] {Perception};
        \node (vehicle) [block, below of=Lplan, yshift=0.5cm] {Vehicle};
        
        \draw [arrow] (Gplan) -- (Lplan);
        \draw [arrow] (map) -- (Gplan);
        \draw [arrow] (sensors) -- (Lplan);
        \draw [arrow] (Lplan) -- (c_sys);
        \draw [arrow] (c_sys) |- (vehicle);
        \draw [arrow] (vehicle) -| (sensors);
    

    \end{tikzpicture}
}

    \caption{Planning, Perception, and Control Navigation Stack}
    \label{fig:nav_stack}
\end{figure}
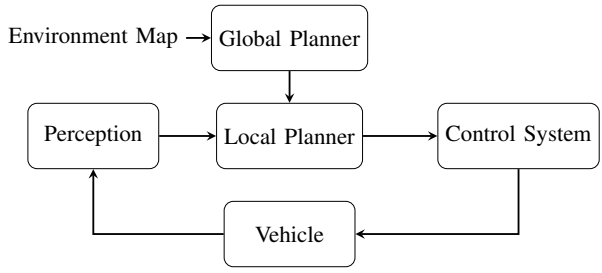

We design our system in the context of a standard perception, planning and control stack as shown in Figure \ref{fig:nav_stack}.
Before the race begins, the global planner uses a map of the environment to generate a reference path of way-points that connect the start and goal locations. The map does not contain any dynamic obstacles. 
During the race, the local planner and control system continuously operates to move the vehicle to the goal location.

\begin{figure}[h]
    \centering
    \def \svgwidth{0.3\textwidth}
\begingroup%
  \makeatletter%
  \providecommand\color[2][]{%
    \errmessage{(Inkscape) Color is used for the text in Inkscape, but the package 'color.sty' is not loaded}%
    \renewcommand\color[2][]{}%
  }%
  \providecommand\transparent[1]{%
    \errmessage{(Inkscape) Transparency is used (non-zero) for the text in Inkscape, but the package 'transparent.sty' is not loaded}%
    \renewcommand\transparent[1]{}%
  }%
  \providecommand\rotatebox[2]{#2}%
  \newcommand*\fsize{\dimexpr\f@size pt\relax}%
  \newcommand*\lineheight[1]{\fontsize{\fsize}{#1\fsize}\selectfont}%
  \ifx\svgwidth\undefined%
    \setlength{\unitlength}{289.82984923bp}%
    \ifx\svgscale\undefined%
      \relax%
    \else%
      \setlength{\unitlength}{\unitlength * \real{\svgscale}}%
    \fi%
  \else%
    \setlength{\unitlength}{\svgwidth}%
  \fi%
  \global\let\svgwidth\undefined%
  \global\let\svgscale\undefined%
  \makeatother%
  \begin{picture}(1,0.66048835)%
    \lineheight{1}%
    \setlength\tabcolsep{0pt}%
    \put(0,0){\includegraphics[width=\unitlength,page=1]{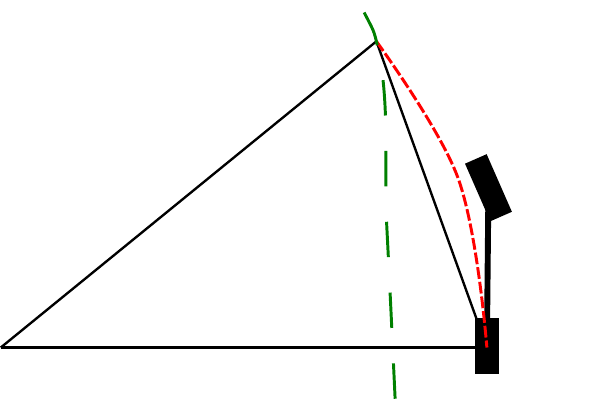}}%
    \put(0.52292344,0.01185513){\color[rgb]{0,0,0}\makebox(0,0)[lt]{\lineheight{1.25}\smash{\begin{tabular}[t]{l}Path\end{tabular}}}}%
    \put(0.83185274,0.20964962){\color[rgb]{0,0,0}\makebox(0,0)[lt]{\lineheight{1.25}\smash{\begin{tabular}[t]{l}L\end{tabular}}}}%
    \put(0.32134162,0.02239348){\color[rgb]{0,0,0}\makebox(0,0)[lt]{\lineheight{1.25}\smash{\begin{tabular}[t]{l}R\end{tabular}}}}%
    \put(0.18690664,0.14284559){\color[rgb]{0,0,0}\makebox(0,0)[lt]{\lineheight{1.25}\smash{\begin{tabular}[t]{l}$2\alpha$\end{tabular}}}}%
    \put(0.70165577,0.54226174){\color[rgb]{0,0,0}\makebox(0,0)[lt]{\lineheight{1.25}\smash{\begin{tabular}[t]{l}$\alpha$\end{tabular}}}}%
    \put(0.65839331,0.27242411){\color[rgb]{0,0,0}\makebox(0,0)[lt]{\lineheight{1.25}\smash{\begin{tabular}[t]{l}$l_\text{d}$\end{tabular}}}}%
    \put(0.26768288,0.36800651){\color[rgb]{0,0,0}\makebox(0,0)[lt]{\lineheight{1.25}\smash{\begin{tabular}[t]{l}R\end{tabular}}}}%
    \put(0,0){\includegraphics[width=\unitlength,page=2]{pp21.pdf}}%
  \end{picture}%
\endgroup%

    \caption{\textbf{Pure Pursuit Path Follower:}$\alpha$ is the angle to the goal, $l_d$ is the look ahead distance, $L$ is the vehicle wheelbase, and $R$ is the turning radius.}
    \label{fig:pure_pursuit}
\end{figure}

We use a pure pursuit path follower which navigates towards a way-point situated at a fixed look-ahead distance in front of the vehicle.
Figure \ref{fig:pure_pursuit} shows how the vehicle and way-points are modelled so that the steering angle can be calculated.

Using the notation from Figure \ref{fig:pure_pursuit}, the desired steering angle, $\delta_\text{pp}$ is calculated according to, 
\begin{equation}
    \delta_\text{pp} = \arctan\left(\frac{2 L \sin(\alpha)}{l_\text{d}}\right).
\end{equation}

The velocity is calculated as the fastest the vehicle can safely travel while turning at the current steering angle.
The velocity is calculated according to the maximum amount of lateral friction the vehicle can withstand, relative to the force exerted by the car in turning based on the current steering angle.
We compute the velocity as,
\begin{equation}
    V = f_\text{s} \sqrt{ \frac{bmg}{ m \tan(|\delta_{\text{ref}|}) /L }},
 \label{eqn:velocity}   
\end{equation}
where $b$ is the coefficient of friction, $m$ is the mass of the vehicle, $L$ is the wheelbase length, and $f_\text{s}$ is a safety factor.
The velocity is limited to the vehicle's maximum velocity.

The navigation stack uses a low-level proportional control system to command the onboard vehicle actuators to follow the local planning references.
The control system calculates acceleration and steering-velocity commands which can be implemented on hardware.

\section{Learning Policies for Reference Modification}\label{sec:learning}
\begin{figure}[h]
    \centering
    \def \svgwidth{0.35\textwidth}
\begingroup%
  \makeatletter%
  \providecommand\color[2][]{%
    \errmessage{(Inkscape) Color is used for the text in Inkscape, but the package 'color.sty' is not loaded}%
    \renewcommand\color[2][]{}%
  }%
  \providecommand\transparent[1]{%
    \errmessage{(Inkscape) Transparency is used (non-zero) for the text in Inkscape, but the package 'transparent.sty' is not loaded}%
    \renewcommand\transparent[1]{}%
  }%
  \providecommand\rotatebox[2]{#2}%
  \newcommand*\fsize{\dimexpr\f@size pt\relax}%
  \newcommand*\lineheight[1]{\fontsize{\fsize}{#1\fsize}\selectfont}%
  \ifx\svgwidth\undefined%
    \setlength{\unitlength}{406.78834088bp}%
    \ifx\svgscale\undefined%
      \relax%
    \else%
      \setlength{\unitlength}{\unitlength * \real{\svgscale}}%
    \fi%
  \else%
    \setlength{\unitlength}{\svgwidth}%
  \fi%
  \global\let\svgwidth\undefined%
  \global\let\svgscale\undefined%
  \makeatother%
  \begin{picture}(1,0.4428635)%
    \lineheight{1}%
    \setlength\tabcolsep{0pt}%
    \put(0,0){\includegraphics[width=\unitlength,page=1]{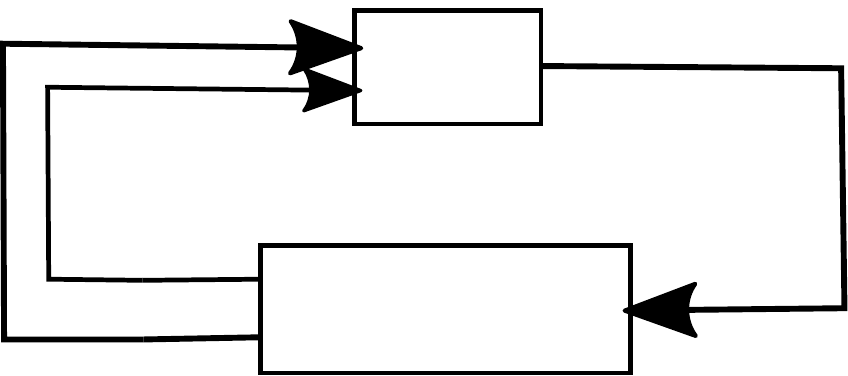}}%
    \put(0.39067677,0.06731728){\color[rgb]{0,0,0}\makebox(0,0)[lt]{\lineheight{1.25}\smash{\begin{tabular}[t]{l}Environment\end{tabular}}}}%
    \put(0.46442517,0.35309241){\color[rgb]{0,0,0}\makebox(0,0)[lt]{\lineheight{1.25}\smash{\begin{tabular}[t]{l}Agent\end{tabular}}}}%
    \put(0.7054245,0.39655128){\color[rgb]{0,0,0}\makebox(0,0)[lt]{\lineheight{1.25}\smash{\begin{tabular}[t]{l}Action\end{tabular}}}}%
    \put(0.11016935,0.28724557){\color[rgb]{0,0,0}\makebox(0,0)[lt]{\lineheight{1.25}\smash{\begin{tabular}[t]{l}Reward\end{tabular}}}}%
    \put(0.11543708,0.41367141){\color[rgb]{0,0,0}\makebox(0,0)[lt]{\lineheight{1.25}\smash{\begin{tabular}[t]{l}State\end{tabular}}}}%
  \end{picture}%
\endgroup%

    \caption{Reinforcement Learning Agent Interacting With Environment}
    \label{fig:general_rl_agent}
\end{figure}
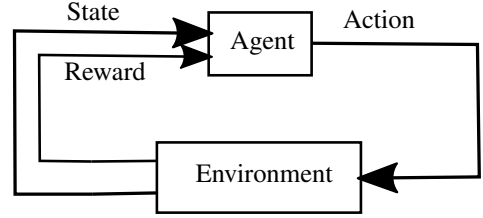

The RL agent comprises a neural network that is trained from experience to maximise a reward signal \cite{sutton2018reinforcement}. 
The neural network is used to represent a policy that maps states to actions.
For each training step, the network receives a state and selects an action that is implemented in the environment as shown in Figure \ref{fig:general_rl_agent}.
The environment then returns the updated state and a reward to the agent and the process is repeated.

We train the agent to modify the pure pursuit steering reference $\delta_{pp}$ to prevent collisions.
The state which the network receives consists of a vector containing the current velocity $V_\text{t}$ and steering of the vehicle $\delta_\text{t}$, the references calculated by the path follower $\delta_{pp}$, and the current range finder scan $[r_\text{1},...,r_\text{10}]$. 
The network outputs an action quantity which is used to modify the steering reference calculated by the path follower.
The path follower (pp) and neural network (nn) steering commands are combined as, 
\begin{equation}
    \delta_\text{ref} =  \delta_\text{pp} + \delta_\text{nn}.
    \label{eqn:ref_mod}
\end{equation}
The reference steering value is limited to be within the maximum range and then sent to the control system to be executed.
The inputs and outputs from the network are in the range $[-1, 1]$ and are scaled according to the maximum values.

\begin{figure}[h]
    \centering
    \def \svgwidth{0.3\textwidth}
\begingroup%
  \makeatletter%
  \providecommand\color[2][]{%
    \errmessage{(Inkscape) Color is used for the text in Inkscape, but the package 'color.sty' is not loaded}%
    \renewcommand\color[2][]{}%
  }%
  \providecommand\transparent[1]{%
    \errmessage{(Inkscape) Transparency is used (non-zero) for the text in Inkscape, but the package 'transparent.sty' is not loaded}%
    \renewcommand\transparent[1]{}%
  }%
  \providecommand\rotatebox[2]{#2}%
  \newcommand*\fsize{\dimexpr\f@size pt\relax}%
  \newcommand*\lineheight[1]{\fontsize{\fsize}{#1\fsize}\selectfont}%
  \ifx\svgwidth\undefined%
    \setlength{\unitlength}{253.99835573bp}%
    \ifx\svgscale\undefined%
      \relax%
    \else%
      \setlength{\unitlength}{\unitlength * \real{\svgscale}}%
    \fi%
  \else%
    \setlength{\unitlength}{\svgwidth}%
  \fi%
  \global\let\svgwidth\undefined%
  \global\let\svgscale\undefined%
  \makeatother%
  \begin{picture}(1,0.58410509)%
    \lineheight{1}%
    \setlength\tabcolsep{0pt}%
    \put(0,0){\includegraphics[width=\unitlength,page=1]{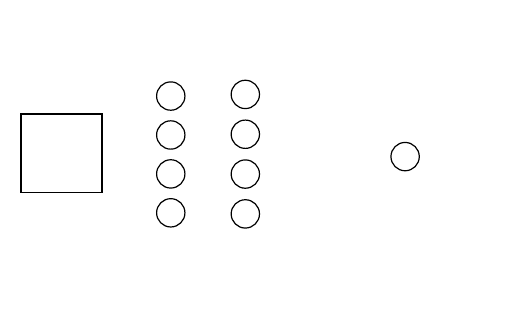}}%
    \put(0.0515623,0.27143163){\color[rgb]{0,0,0}\makebox(0,0)[lt]{\lineheight{1.25}\smash{\begin{tabular}[t]{l}State\end{tabular}}}}%
    \put(0.030583,0.14568533){\color[rgb]{0,0,0}\makebox(0,0)[lt]{\lineheight{1.25}\smash{\begin{tabular}[t]{l}$\delta_\text{pp},\delta_\text{t}, V$\end{tabular}}}}%
    \put(0,0){\includegraphics[width=\unitlength,page=2]{nn_arch1.pdf}}%
    \put(0.80806962,0.31040631){\color[rgb]{0,0,0}\makebox(0,0)[lt]{\lineheight{1.25}\smash{\begin{tabular}[t]{l}Action\\$\delta_\text{nn}$\end{tabular}}}}%
    \put(0,0){\includegraphics[width=\unitlength,page=3]{nn_arch1.pdf}}%
    \put(0.00,0.38347465){\color[rgb]{0,0,0}\makebox(0,0)[lt]{\lineheight{1.25}\smash{\begin{tabular}[t]{l}$r_\text{1},...,r_\text{10}$\end{tabular}}}}%
  \end{picture}%
\endgroup%

    \caption{\textbf{Reinforcement Learning Agent Architecture:} The state consists of 10 sparse laser range finders and the pure pursuit steering reference. A neural network with two fully connected layers is used to calculate an action which is the steering reference which is used to modify the path follower reference.}
    \label{fig:rl_agent}
\end{figure}


The aim in training the network is that the vehicle does not crash and that it tracks the reference path closely.
Therefore, we use a reward signal which severely punishes crashing and rewards progress along the reference path.
We write our reward signal as,
\begin{equation}
    r(\textbf{s}_\text{t}, \textbf{a}_\text{t}) = 
    \begin{cases}
        r_{\text{crash}} = -1 & \quad  \text{if crash} \\
        r_\text{complete} = 1 & \quad  \text{if lap complete} \\
        r_\text{tracking} = (s_\text{t+1} - s_\text{t}) / s_\text{total} & \quad \text{otherwise},
    \end{cases}
\end{equation}
where $s$ is the progress along the track at a specific step.
The sum of the tracking rewards for completing the episode will always be 1 since the sum of the progress differences will be equal to the track length.
The tracking reward must be a potential function of the terminal reward so that the agent is rewarded in proportion to the progress towards the ideal terminal state.

\section{Evaluation Methodology}\label{sec:evaluation_method}

The ability of the modification planner to follow a reference path and avoid obstacles efficiently is evaluated in the context of F1/$10^{th}$ autonomous racing.
We aim to determine the performance of the modification planner and compare it to current baseline solutions.
We measure performance in terms of the following metrics, i.e.  (1) the time to navigate to a target and (2) the success rate, which is the percentage of episodes successfully completed without crashing.

\begin{figure}[h]
    \centering
    \def \svgwidth{0.26\textwidth}
    \input{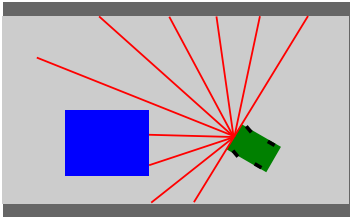}
    \caption{F1/$10^{th}$ Vehicle Simulator with 10 Laser Range Finder Beams }
    \label{fig:simulator}
\end{figure}

\subsection{Simulation Environment}

We train and test our local planner architectures using a custom-built, F1/$10^{th}$ autonomous racing simulator.
The simulator models a non-holonomic 1/$10^{th}$ scale autonomous race car with a LiDAR sensor as shown in Figure \ref{fig:simulator}.
At each time-step, the simulator receives an action, updates the state according to the transition dynamics and then returns the new state of the vehicle.
A version of our test code is available online
\footnote{Available at: https://github.com/BDEvan5/ReferenceModification}.

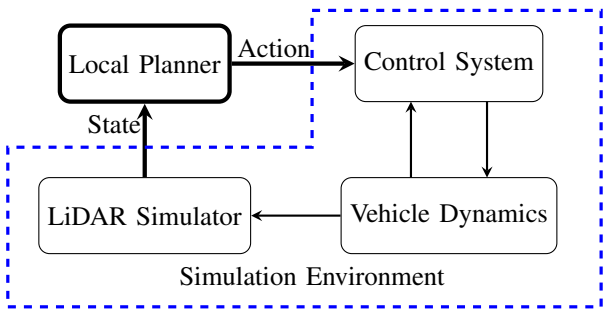
\begin{figure}[h]
    \centering
    \tikzstyle{empty} = [rectangle, rounded corners, draw=white, minimum height = 1cm, minimum width=2cm]
\tikzstyle{block_t} = [rectangle, ultra thick, rounded corners, draw=black, minimum height = 1cm, minimum width=2cm]
\tikzstyle{block} = [rectangle, rounded corners, draw=black, minimum height = 1cm, minimum width=2cm]
\tikzstyle{dash_block} = [rectangle, square corners, draw=black, ]

\tikzstyle{arrow} = [thick, ->, >=stealth]
\tikzstyle{arrow_t} = [ultra thick, ->, >=stealth]
\tikzstyle{ddash} = [dashed, very thick, blue]




\resizebox{0.46\textwidth}{!}{
    \begin{tikzpicture}[node distance=2cm]
    
        \node (lplan) [block_t] {Local Planner};
        \node (csys) [block, right of=lplan, xshift=2cm] {Control System};
        \node (lidar) [block, below of=lplan] {LiDAR Simulator};
        \node (dyn) [block, below of=csys] {Vehicle Dynamics};
        
        \draw [arrow_t] (lplan) -- (csys);
        \draw [arrow] (4.5, -0.5) -- (4.5, -1.5);
        \draw [arrow] (3.5, -1.5) -- (3.5, -0.5);
        \draw [arrow] (dyn) -- (lidar);
        \draw [arrow_t] (lidar) -- (lplan);
        
    
        \draw[ddash] (2.2, 0.7) -- (6, 0.7) -- (6, -3.2) -- (-1.8, -3.2) -- (-1.8, -1.1) -- (2.2, -1.1) -- (2.2, 0.7);
        
        \node[align=left] at (1.7, 0.2) {Action};
        \node[align=left] at (-0.4, -0.8) {State};
        \node[align=center] at (2.2, -2.8) {Simulation Environment};
    
    \end{tikzpicture}
}
    \caption{Schematic of Planning Loop in Simulation Environment}
    \label{fig:sim_env}
\end{figure}

The car is modelled using the kinematic bicycle model \cite{kinematic_bicycle_model}, with the vehicle state consisting of, position, orientation, velocity, and steering angle.
Figure \ref{fig:sim_env} shows how the simulator interfaces with the local planner that is being evaluated.
The local planner selects an action consisting of velocity and steering angle commands which are passed to the simulator and used as the references for the control system.
The feedback control system calculates acceleration and steering-velocity commands which are the input to the dynamics equations.
As in practice, the simulated control system runs at a higher frequency of 100Hz compared to the local planner which runs at 10Hz.
Table \ref{tab:sim_params} notes several important parameters used in our simulation.

\begin{table}[h]
    \centering
    \begin{tabular}{p{5cm}|m{1.5cm}}
        \textbf{Simulation Parameter} & \textbf{Value} \\
        \hline
        Simulation dynamics time step & 0.01 s  \\ 
        Local planning time step & 0.1 s \\ 
        Max range finder value & 10 m \\
        Range finder noise $\sigma_{s}$ & 0.01 \\
        Max vehicle speed & 7 m/s \\
        Max steering angle & 0.4 rad \\
        Pure Pursuit Look-ahead distance & 0.8 m \\
    \end{tabular}
    \caption{Parameters Used in the Simulation}
    \label{tab:sim_params}
\end{table}

The new vehicle state is used to simulate a LiDAR scan of 10 sparse range finders on the map and the new state and the scan is returned to the local planner.
To simulate the noise in range finder readings, Gaussian distributed process noise, $\mathcal{N}(0, \sigma_{s})$, is added to the measurements.

\subsection{Baseline Solutions}

We compare the modification planner to three common baseline local planners.
The first baseline is the ``Follow the Gap Method'' (FGM) \cite{sezer2012novel}.
The FGM method identifies a bubble around the nearest obstacle and then navigates away from the bubble into free space.
In order for the FGM method to accurately identify a gap, it requires a dense LiDAR scan of many beams.
We thus we adapt the simulator to produce 1080 beams.

The second baseline which is used is an optimisation based, oracle planner which has full access to a map of the environment and the location of the obstacles before the episode begins.
The obstacle map is used to generate an optimal path around the obstacles.
The optimal plan is followed using the pure pursuit algorithm.
This solution is treated as the best performance which could be achieved by any local planner.

The main baseline that we compare our solution to is a end-to-end planner that uses an RL agent to replace the entire navigation pipeline.
The end-to-end planner uses the relative target location and laser range finder readings to calculate a steering angle that leads the vehicle around obstacles towards the target.
The end-to-end planner is trained in the same way as the modification planner using the same neural network parameters and reward signal.

For all of our baseline implementations, we use the same method of calculating the speed based on the steering angle to ensure that the results are comparable.

\subsection{Reinforcement Learning}

\begin{figure}[h]
    \centering


\begin{tikzpicture}
	\begin{axis}
		[name=plot,width=.48\textwidth,
		height=4cm,
		xmin=-50, xmax=4000,
		ymin=-0.8, ymax=2,
		grid=major,
		grid style=dashed,
		xlabel={Episode Number},
		ylabel={Average Reward}]

		\pgfplotstableread[col sep=comma,]{Graphs/TrainingData.csv}{\table}
		
		\addplot[color=blue, ultra thick] table[x={x}, y={y}]{\table};
		\label{training_plot}
	
	\end{axis}
\end{tikzpicture}

    \caption{Modification Planner Training Graph}
    \label{fig:training_graph}
\end{figure}
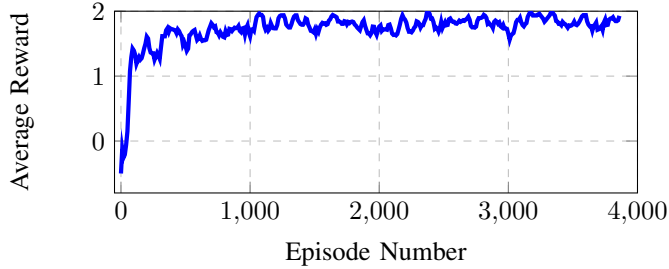

The RL agent in the modification planner is trained using the Twin Delayed Deterministic Policy Gradient (TD3) reinforcement learning algorithm with the original training hyper-parameters \cite{td3_paper}.
The TD3 algorithm is selected since it is a current state of the art algorithm in many RL challenges.
Neural networks with two hidden layers of 200 neurons are trained in mini-batches of 100 transitions.
The training lasts for 200,000 steps which is generally just below 4000 episodes.
200,000 steps was selected after evaluating the agent after different numbers of steps to ensure and seeing that it had always converged by 200,000.
Figure \ref{fig:training_graph} shows the average reward per episode for the training of the network.
The network converges towards a value of 2 which is made up from the total of the tracking reward and terminal reward.

\section{Evaluation Results}\label{sec:results}

\subsection{Random Forests}
The modification planner is evaluated in a random forest which consists of a straight race track with randomly located obstacles.
The random forest is a straight track (20m x 2m) where 4 obstacles (0.5m x 0.5m) are randomly placed every episode along the track. 
Table \ref{tab:eval_results_forest} shows the results from running 100 test episodes and taking the average times to navigate to goal and success rate.

\begin{table}[h]
\renewcommand{\arraystretch}{1.5}
        \begin{tabular}{m{0.15\textwidth}|m{0.1\textwidth}|m{0.08\textwidth} c}
        \textbf{Vehicle}  & \textbf{No Obstacles} & \textbf{Obstacles} & \textbf{Success} \\
         \hline
            Follow the Gap Method & 3.73 & 4.38 & 99\% \\
            Oracle & 3.72 & 3.87 & 100\% \\
            End-to-end Planner & 5.08 & 6.18 & 97\% \\
            Modification Planner & 3.82 & 4.63 & 94\% 
    \end{tabular}
    \caption{Results from 20m Random Forest}
    \label{tab:eval_results_forest}
\end{table}
In the tests with no obstacles the modification planner is able to track the optimal path, straight line through the forest, and achieve a very similar time to the optimisation based oracle planner (within 3\% of the oracle planner).
The similar times to the optimisation based planner when no obstacles are present shows that the RL agent learns to not modify the steering command unnecessarily.
The end-to-end planner achieves a significantly slower average time of 5.08 seconds in this test compared to the 3.72 seconds which the other planners are able to achieve.

\begin{figure}[h]
    \centering
    \def \svgwidth{0.43\textwidth}
\begingroup%
  \makeatletter%
  \providecommand\color[2][]{%
    \errmessage{(Inkscape) Color is used for the text in Inkscape, but the package 'color.sty' is not loaded}%
    \renewcommand\color[2][]{}%
  }%
  \providecommand\transparent[1]{%
    \errmessage{(Inkscape) Transparency is used (non-zero) for the text in Inkscape, but the package 'transparent.sty' is not loaded}%
    \renewcommand\transparent[1]{}%
  }%
  \providecommand\rotatebox[2]{#2}%
  \newcommand*\fsize{\dimexpr\f@size pt\relax}%
  \newcommand*\lineheight[1]{\fontsize{\fsize}{#1\fsize}\selectfont}%
  \ifx\svgwidth\undefined%
    \setlength{\unitlength}{278.61825757bp}%
    \ifx\svgscale\undefined%
      \relax%
    \else%
      \setlength{\unitlength}{\unitlength * \real{\svgscale}}%
    \fi%
  \else%
    \setlength{\unitlength}{\svgwidth}%
  \fi%
  \global\let\svgwidth\undefined%
  \global\let\svgscale\undefined%
  \makeatother%
  \begin{picture}(1,0.61833323)%
    \lineheight{1}%
    \setlength\tabcolsep{0pt}%
    \put(0,0){\includegraphics[width=\unitlength,page=1]{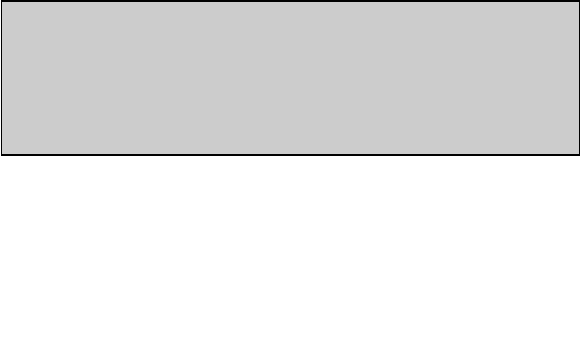}}%
    \put(0.04286529,0.47067798){\color[rgb]{0,0,0}\makebox(0,0)[lt]{\lineheight{1.25}\smash{\begin{tabular}[t]{l}x\end{tabular}}}}%
    \put(0.93929313,0.47324157){\color[rgb]{0,0,0}\makebox(0,0)[lt]{\lineheight{1.25}\smash{\begin{tabular}[t]{l}x\end{tabular}}}}%
    \put(0,0){\includegraphics[width=\unitlength,page=2]{nav_mod_comp1.pdf}}%
    \put(0.04190392,0.12073686){\color[rgb]{0,0,0}\makebox(0,0)[lt]{\lineheight{1.25}\smash{\begin{tabular}[t]{l}x\end{tabular}}}}%
    \put(0.93833172,0.12330045){\color[rgb]{0,0,0}\makebox(0,0)[lt]{\lineheight{1.25}\smash{\begin{tabular}[t]{l}x\end{tabular}}}}%
    \put(0,0){\includegraphics[width=\unitlength,page=3]{nav_mod_comp1.pdf}}%
    \put(0.06113142,0.376463){\color[rgb]{0,0,0}\makebox(0,0)[lt]{\lineheight{1.25}\smash{\begin{tabular}[t]{l}Modification\end{tabular}}}}%
    \put(0.07170662,0.02363764){\color[rgb]{0,0,0}\makebox(0,0)[lt]{\lineheight{1.25}\smash{\begin{tabular}[t]{l}Navigation\end{tabular}}}}%
  \end{picture}%
\endgroup%

    \caption{Example Trajectories Executed by the Modification and End-to-end Planners}
    \label{fig:mod_nav_comp}
\end{figure}

In the tests with obstacles present, the optimisation based oracle planner outperforms the other planners as expected with an average time of 3.87 seconds.
The FGM planner achieves a mediocre time of 4.33 seconds, however, this is still faster than any of the learning-based planners.
Reactive methods perform well in highly structured environments where the best route is straight forward, and complex velocity profiles are not required.
The modification planner achieves an average time of 4.63 seconds for the forest filled with obstacles which is faster than the end-to-end planners time of 6.18 seconds. 
The decrease in time shows that the modification planner is able to find smoother, faster trajectories than the end-to-end solution.
Figure \ref{fig:mod_nav_comp} shows a comparison of the trajectories taken by the end-to-end and modification planners and how the modification planner learns to take smoother trajectories.

The modification planner achieves a slightly lower success rate of 94\% compared to the end-to-end planner's 97\%. 
These high success rates show that both planners learn to avoid a significant majority of obstacles.
Many similar papers in the field of autonomous racing show the similar result of learning to race, yet still occasionally crashing \cite{perot2017end, vitelli2016carma}.

The influence of the network in modifying the path follower references is shown below.
Figure \ref{fig:mod_nn} shows a graph of the output from the neural network as the vehicle moves along the path.
The neural network modifies the path enough to avoid the obstacles while roughly maintaining the reference path.

\begin{figure}[h]
    \centering
    \def \svgwidth{0.43\textwidth}
\begingroup%
  \makeatletter%
  \providecommand\color[2][]{%
    \errmessage{(Inkscape) Color is used for the text in Inkscape, but the package 'color.sty' is not loaded}%
    \renewcommand\color[2][]{}%
  }%
  \providecommand\transparent[1]{%
    \errmessage{(Inkscape) Transparency is used (non-zero) for the text in Inkscape, but the package 'transparent.sty' is not loaded}%
    \renewcommand\transparent[1]{}%
  }%
  \providecommand\rotatebox[2]{#2}%
  \newcommand*\fsize{\dimexpr\f@size pt\relax}%
  \newcommand*\lineheight[1]{\fontsize{\fsize}{#1\fsize}\selectfont}%
  \ifx\svgwidth\undefined%
    \setlength{\unitlength}{302.67856267bp}%
    \ifx\svgscale\undefined%
      \relax%
    \else%
      \setlength{\unitlength}{\unitlength * \real{\svgscale}}%
    \fi%
  \else%
    \setlength{\unitlength}{\svgwidth}%
  \fi%
  \global\let\svgwidth\undefined%
  \global\let\svgscale\undefined%
  \makeatother%
  \begin{picture}(1,0.28672575)%
    \lineheight{1}%
    \setlength\tabcolsep{0pt}%
    \put(0,0){\includegraphics[width=\unitlength,page=1]{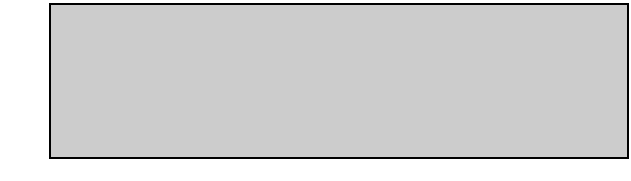}}%
    \put(0.11608846,0.17110125){\color[rgb]{0,0,0}\makebox(0,0)[lt]{\lineheight{1.25}\smash{\begin{tabular}[t]{l}x\end{tabular}}}}%
    \put(0.94125811,0.16874145){\color[rgb]{0,0,0}\makebox(0,0)[lt]{\lineheight{1.25}\smash{\begin{tabular}[t]{l}x\end{tabular}}}}%
    \put(0,0){\includegraphics[width=\unitlength,page=2]{mod_nn_out1.pdf}}%
  \end{picture}%
\endgroup%

    \begin{tikzpicture}
	\begin{axis}
		[name=plot,width=.48\textwidth, height=4cm,
		xmin=0, xmax=20,
		ymin=-0.5, ymax=0.5,
		grid=major,
		grid style=dashed,
		xlabel={Position}]
		
		\addplot[color=blue, ultra thick] table[col sep=comma]{Graphs/nn_output.csv};
		\label{training_plot}
	
	\end{axis}
\end{tikzpicture}
    
    \caption{Example Trajectory from the Modification Planner with Neural Network Output. The y-axis label (hidden due to size) is the scaled output from the neural network.}
    \label{fig:mod_nn}
\end{figure}

\subsection{Repeatability}

To measure the repeatability of the learned policies, we train and evaluate 10 agents on the random forest described above.
The average and standard deviation statistics are presented in Table \ref{tab:repeat}.

\begin{table}[h]
    \centering
    \begin{tabular}{m{3.8cm}|c|c}
        \textbf{Statistic} & \textbf{Average} & \textbf{Standard Deviation} \\
        \hline
        No Obstacle Average Time & 3.85 s & $\pm$ 0.10 \\
        Obstacle Average Time & 4.63 s & $\pm$ 0.17 \\
        Obstacle Success Rate & 92.2\% & $\pm$ 5.3 \\
    \end{tabular}
    \caption{Results from Repeatability Evaluation of the Modification Planner}
    \label{tab:repeat}
\end{table}

The standard deviation in the times achieved by the different agents is small and this shows that the method of training agents to achieve similar times is stable and highly repeatable.
The standard deviation of the success rate is large, 5\%, which shows that there is a non-trivial difference in how agents learn to avoid obstacles.

\subsection{Discussion}

The results show that the modification planner can learn a policy that produces faster performance than the end-to-end planner.
Specifically, the modification planner avoids obstacles more smoothly which results in the vehicle being able to travel at higher speeds.
This shows that the modification planner achieves the behaviour that it was designed for, allowing non-holonomic vehicles to avoid obstacles smoothly at high speeds.
The success rate which the planners achieve show that while the modification planner is able to learn to avoid a significant number of obstacles, it still underperforms the end-to-end planners ability to consistently avoid obstacles.

It is suggested that the reason for the lower rate of completion is that the complexity of the learning problem for the modification planner is higher than for the end-to-end baseline.
The transition dynamics (movement from one state to the next) for the modification planner are dependant on combining the action, which the agent has selected, with the pure pursuit reference. 
The result of this combination is that the agent has less influence on how the vehicle moves.
Having less influence improves the quality of the path taken which results in the modification planner achieving faster times.
A problem that arises in having less influence is that it achieves a lower rate of completion.

This problem of guaranteeing vehicle safety is a current topic of research and could be solved in future work by exploring different training methods such as imitation learning.
Another possible solution is to combine the RL agent with a safety system that can stop the vehicle before it crashes.

A recent survey \cite{xiao2020motion} suggested that future work in mobile navigation should continue into using ML to learn specific subsystems of the navigation pipeline. 
In future, we plan to investigate how RL agents can be used in different architectures to learn obstacle avoidance behaviours by reducing the complexity of the navigation problem to more simple components.

\section{Conclusion}
In this paper, we addressed the problem of autonomous racing without an explicit obstacle map.
We presented a novel planning architecture that uses a path following algorithm in conjunction with an RL agent to smoothly avoid obstacles while maintaining the global plan.
Our architecture is evaluated in the context of F1/$10^{th}$ autonomous racing and shown to achieve comparable racing times to optimisation based and reactive methods, yet using fewer inputs.
Specifically, our modification local planner requires only 10 sparse laser range finder readings and is able to smoothly avoid obstacles.
We showed that our planner is able to avoid obstacles more smoothly than previous solutions that were trained end-to-end.

\addtolength{\textheight}{-12cm}   

\typeout{}


\end{document}